\documentclass[conference]{IEEEtran}
\IEEEoverridecommandlockouts

\usepackage{cite}
\usepackage{amsmath,amssymb,amsfonts}
\usepackage{algorithmic}
\usepackage{algorithm}
\usepackage{graphicx}
\usepackage{textcomp}
\usepackage{xcolor}
\usepackage{booktabs}
\usepackage{multirow}
\usepackage{url}
\usepackage{amsthm}

\newtheorem{remark}{Remark}

\graphicspath{{figs_paper/}}

\begin{document}

\title{Smoother Action Chunking Flow Policy via\\
Prior-Corrected Orthogonal Trust-Region Guidance}

\author{
\IEEEauthorblockN{1\textsuperscript{st} Kai Fang}
\IEEEauthorblockA{\textit{South China University of Technology}\\
Guangzhou, China\\
thissfk@qq.com}
\and
\IEEEauthorblockN{2\textsuperscript{nd} Hailong Pei}
\IEEEauthorblockA{\textit{South China University of Technology}\\
Guangzhou, China\\
auhlpei@scut.edu.cn}
\and
\IEEEauthorblockN{3\textsuperscript{rd} Xuemin Chi}
\IEEEauthorblockA{\textit{Zhejiang University}\\
Hangzhou, China\\
xueminchisnow@gmail.com}
}

\maketitle

\begin{abstract}
Flow-matching robot policies commonly use action-chunking inference for efficient closed-loop control, but chunk boundaries can introduce discontinuous action transitions. Existing RTC guidance improves continuity by injecting correction signals during denoising, yet its weight schedule is weak at intermediate timesteps and its unconstrained correction direction may introduce transverse perturbations. We propose POTR, a \textbf{p}rior-corrected \textbf{o}rthogonal \textbf{t}rust-\textbf{r}egion guidance method. First, we incorporate a data-prior scale $\sigma_d$ into the RTC guidance weight, yielding stronger intermediate-time correction. Second, we decompose the guidance vector into components parallel and perpendicular to the denoising velocity, and constrain the perpendicular component within a trust region. On LIBERO with $\pi_{0.5}$, POTR improves success rate and consistently reduces chunk-boundary discontinuity, acceleration, and jerk compared with RTC. Ablations show that the prior-corrected weight provides the main correction gain, while the orthogonal trust region further improves stability.
\end{abstract}


\section{Introduction}

Flow matching \cite{b1,b2} enables robot manipulation policy learning with straighter sampling trajectories than DDPM \cite{b3}. $\pi_0$ \cite{b4} applies flow matching to VLA models; Diffusion Policy \cite{b5} combines DDPM with chunked prediction. Asynchronous action-chunking inference \cite{b6} improves efficiency but introduces L2 jumps at chunk boundaries, causing acceleration and jerk spikes that degrade safety \cite{b7}.

The Real-Time Chunking (RTC) method \cite{b8} injects observation-guided correction during denoising, building on $\Pi\text{GDM}$ \cite{b9} and inpainting \cite{b10}. The EDM framework \cite{b11} provides a theoretical basis for data scale in guidance. However, RTC has two limitations: (1) the weight at $\tau \approx 0.5$ is only $\sim$2.0, leaving intermediate denoising under-corrected; (2) the correction direction is unconstrained, producing perpendicular perturbations at chunk boundaries. Alternatives like BID \cite{b12} and Streaming Diffusion Policy \cite{b13} require architectural changes, whereas our method modifies only the guidance signal.

We propose POTR with two components: (1) a \textbf{prior-corrected weight} using $\sigma_d$ to boost intermediate guidance by $\sim$$3.6\times$; (2) an \textbf{orthogonal trust-region constraint} clipping the perpendicular component ($\|\mathbf{g}_\perp'\| \leq \rho\|\mathbf{v}_\tau\|$). We evaluate on LIBERO \cite{b14} with 6 metrics spanning success rate, efficiency, and smoothness.

\section{Preliminaries and Motivation}

\subsection{Flow Matching and Action Chunking}

Consider a flow-matching policy $\pi_\theta$ with asynchronous chunked inference: every $s$ steps the model generates an action chunk $\mathbf{A}_t = [a_t, \ldots, a_{t+H-1}]$ of length $H$. The conditional flow matching path is:
\begin{equation}
p_\tau(\mathbf{A}_t^\tau \mid \mathbf{A}_t^1) = \mathcal{N}\!\left(\tau \mathbf{A}_t^1,\; (1-\tau)^2 \mathbf{I}\right)
\label{eq:1}
\end{equation}
where $\tau \in [0, 1]$ is the denoising timestep. Discretization via $n$-step Euler solver:
\begin{equation}
\mathbf{A}_t^{\tau + 1/n} = \mathbf{A}_t^\tau + \frac{1}{n}\, v_\pi(\mathbf{A}_t^\tau, \mathbf{o}_t, \tau)
\label{eq:2}
\end{equation}

\subsection{RTC Guidance Framework}

RTC \cite{b8} injects residual actions from the previous chunk as an inpainting \cite{b10} target at each denoising step, building on $\Pi\text{GDM}$ \cite{b9}:
\begin{multline}
\mathbf{v}_{\Pi\text{GDM}} = \mathbf{v}_\pi + \min\!\left(\beta,\; \frac{1-\tau}{\tau \cdot r_\tau^2}\right) \cdot (\mathbf{Y} - \hat{\mathbf{A}}_t^1)^\top \text{diag}(\mathbf{W})\, \frac{\partial \hat{\mathbf{A}}_t^1}{\partial \mathbf{A}_t^\tau}
\label{eq:3}
\end{multline}
where $\hat{\mathbf{A}}_t^1 = \mathbf{A}_t^\tau + (1-\tau)\,\mathbf{v}_\pi$ is the one-step clean estimate, $\mathbf{Y}$ is the inpainting target (residual actions from previous chunk), $\mathbf{W} \in \mathbb{R}^H$ is the soft mask, $r_\tau^2 = (1-\tau)^2 / (\tau^2 + (1-\tau)^2)$, and $\beta$ is the clipping threshold. The guidance weight is $w_{\text{RTC}}(\tau) = \min((1-\tau)/(\tau \cdot r_\tau^2),\; \beta)$. Defining $\mathrm{SNR}(\tau) = \tau^2/(1-\tau)^2$, this expands to $w_{\text{RTC}}(\tau) = \min((1-\tau)(1 + \mathrm{SNR}(\tau))/\tau,\; \beta)$.

The weight forms a bimodal structure: at $\tau \to 0$ and $\tau \to 1$ it is high, but at the midpoint $\tau \approx 0.5$ it reaches its minimum ($r^{-2} = 2$) --- precisely where the intermediate denoising phase often plays an important role in shaping the generated action structure. This mismatch is the primary limitation of RTC.

\section{Method}

\subsection{Prior-Corrected Weight Function}

\begin{figure}[htbp]
\centering
\includegraphics[width=\columnwidth]{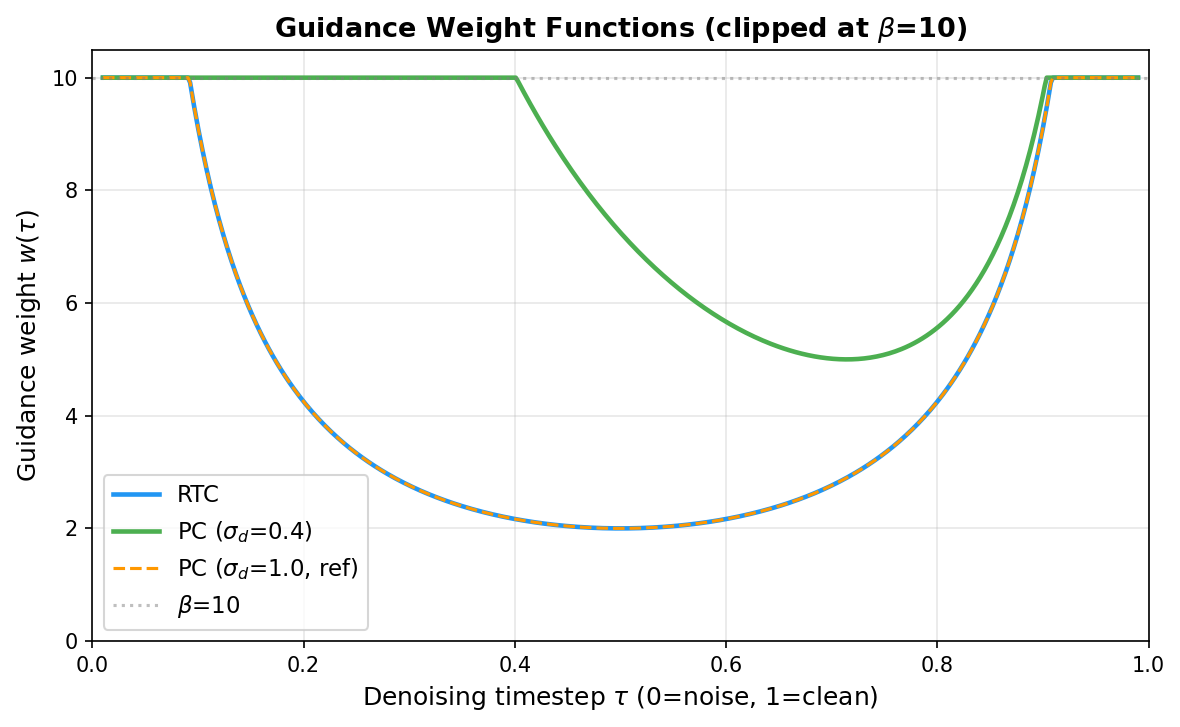}
\caption{Guidance weight function comparison. The prior-corrected weight ($\sigma_d$=0.4) is substantially higher than RTC at intermediate timesteps ($\tau$=0.4--0.7), strengthening correction during the middle denoising phase. At $\sigma_d$=1.0, the formula reduces to a profile close to RTC.}
\label{fig:1}
\end{figure}

RTC and $\Pi\text{GDM}$ \cite{b9} default to $\sigma_d = 1$ (unit prior variance), assuming pixel-normalized image generation. In robot action generation, the observation-conditioned prior $p_1(\mathbf{A}_t^1 \mid \mathbf{o}_t)$ has much smaller variance ($\sigma_{d|\mathbf{o}}^2 \ll 1$): given a current observation, plausible trajectories are highly concentrated. Retaining $\sigma_d$ in the weight derivation gives the corrected normalization:
\begin{equation}
r_\tau^2 = \frac{(1-\tau)^2 \sigma_d^2}{(1-\tau)^2 + \sigma_d^2 \tau^2}
\label{eq:11}
\end{equation}
which reduces to RTC's $r_\tau^2 = (1-\tau)^2/(\tau^2 + (1-\tau)^2)$ at $\sigma_d = 1$. The prior-corrected weight is:
\begin{equation}
w_{\text{PC}}(\tau) = \min\!\left(\frac{(1-\tau)^2 + \sigma_d^2 \tau^2}{\sigma_d^2 \tau (1-\tau)},\; \beta\right)
\label{eq:12}
\end{equation}
With $\sigma_d = 0.4$ at $\tau = 0.5$: $w_\text{PC}(0.5) = 7.25$ vs.\ RTC's $w_\text{RTC}(0.5) = 2.0$, a \textbf{$3.6\times$} increase. Table~\ref{tab:1} and Fig.~\ref{fig:1} show the comparison.

\setcounter{table}{0}
\begin{table}[htbp]
\caption{Guidance Weight Comparison at Key Timesteps ($\sigma_d = 0.4$, $\beta = 10$)}
\label{tab:1}
\centering
\begin{tabular}{|c|c|c|c|}
\hline
$\tau$ & $w_{\text{RTC}}$ & $w_{\text{PC}}$ (clipped) & $w_{\text{PC}} / w_{\text{RTC}}$ \\
\hline
0.1 & 9.11 & 10.00 & $1.10\times$ \\
0.3 & 2.76 & 10.00 & $3.62\times$ \\
0.5 & 2.00 & 7.25 & $3.63\times$ \\
0.7 & 2.76 & 5.01 & $1.82\times$ \\
0.9 & 9.11 & 9.69 & $1.06\times$ \\
\hline
\end{tabular}
\end{table}

\subsection{Orthogonal Trust-Region Constraint}

\begin{figure}[htbp]
\centering
\includegraphics[width=\columnwidth]{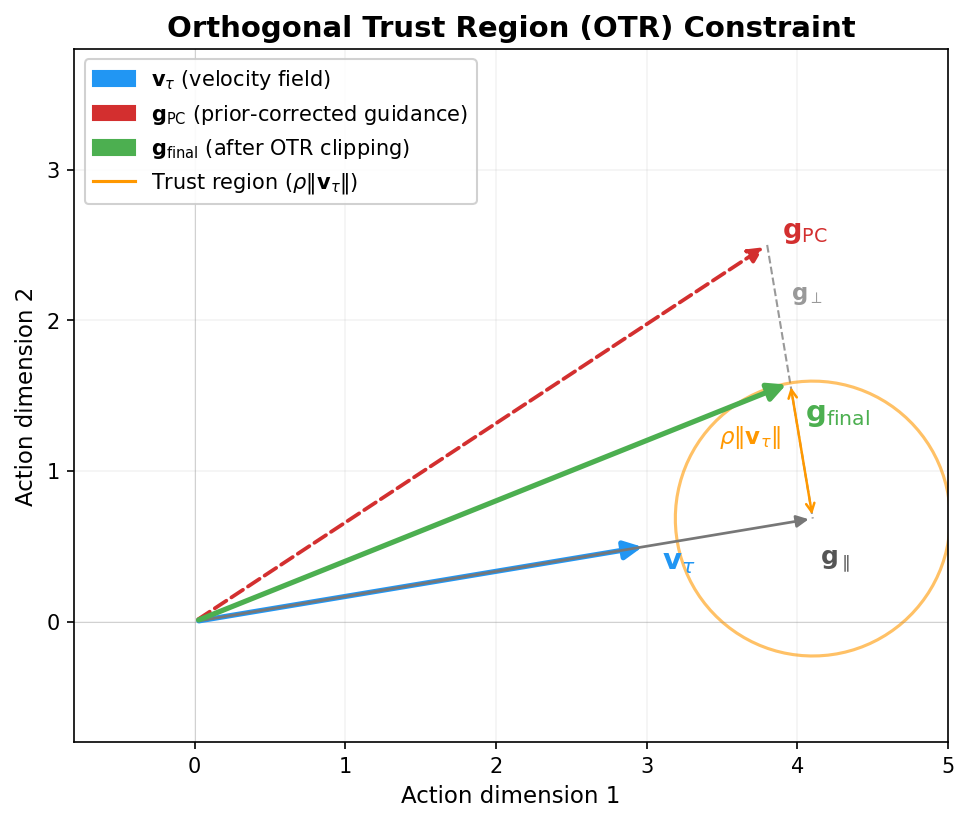}
\caption{Illustration of the OTR constraint. $\mathbf{g}_\text{PC}$ (red dashed) is the prior-corrected guidance vector, decomposed into a parallel component $\mathbf{g}_\parallel$ along $\mathbf{v}_\tau$ (blue) and a perpendicular component $\mathbf{g}_\perp$. The trust region (orange circle) is centered at $\mathbf{g}_\parallel$ with radius $\rho\|\mathbf{v}_\tau\|$. The final guidance $\mathbf{g}_\text{final}$ (green) retains the full parallel component while the perpendicular component is clipped to the trust-region boundary.}
\label{fig:2}
\end{figure}

Boosting guidance weights amplifies both parallel and perpendicular components. The perpendicular component causes abrupt direction changes at chunk boundaries. We constrain it via:
\begin{subequations}
\begin{align}
\max_{\hat{\mathbf{g}}} \quad & \langle \hat{\mathbf{g}},\, \mathbf{g}_{\text{PC}} \rangle \label{eq:13a}\\
\text{s.t.} \quad & \bigl\|\hat{\mathbf{g}} - \mathbf{g}_\parallel\bigr\|_2 \leq \rho\, \|\mathbf{v}_\tau\|_2 \label{eq:13b}
\end{align}
\end{subequations}
where $\mathbf{g}_\parallel = \frac{\mathbf{g}_{\text{PC}} \cdot \mathbf{v}_\tau}{\|\mathbf{v}_\tau\|^2}\, \mathbf{v}_\tau$ and $\mathbf{g}_\perp = \mathbf{g}_{\text{PC}} - \mathbf{g}_\parallel$. The closed-form solution is:
\begin{equation}
\mathbf{g}_{\text{final}} = \mathbf{g}_\parallel + \min\!\left(\frac{\rho \|\mathbf{v}_\tau\|}{\|\mathbf{g}_\perp\|},\; 1\right) \mathbf{g}_\perp
\label{eq:16}
\end{equation}
which bounds the perpendicular component by $\rho\|\mathbf{v}_\tau\|$. Fig.~\ref{fig:2} illustrates the constraint.

\begin{remark}[Synergy between prior correction and OTR]
The two improvements are orthogonally complementary: the prior-corrected weight uniformly amplifies guidance strength along the correction direction, while the orthogonal trust-region constraint selectively clips the perpendicular component exceeding the trust region. Using prior-corrected weights alone amplifies the perpendicular component, exacerbating directional perturbation (risky); using the orthogonal trust region alone is limited in effectiveness when the original weight is already low; only their combination achieves both strong guidance and directional stability. The ablation study (Section~\ref{sec:ablation}) quantitatively validates this synergy across all 6 metrics.
\end{remark}

\subsection{Complete Algorithm}

Algorithm~\ref{alg:pcot} presents the complete implementation of the proposed method within an Euler denoising solver.

\begin{algorithm*}[!t]
\caption{POTR Guided Denoising: \textbf{P}rior-corrected \textbf{O}rthogonal \textbf{T}rust-\textbf{R}egion (Euler Solver)}
\label{alg:pcot}
\begin{algorithmic}[1]
\REQUIRE Trained velocity field model $v_\theta$; initial noise $\mathbf{A}^0 \sim \mathcal{N}(0, \mathbf{I})$; observation $\mathbf{o}$; inpainting target $\mathbf{Y}$ (residual actions from previous chunk); soft mask $\mathbf{W}$; parameters $\sigma_d$, $\rho$, $\beta$, denoising steps $n$
\ENSURE Clean action chunk $\mathbf{A}^1$
\FOR{$k = 0, 1, \ldots, n-1$}
  \STATE $\tau \leftarrow k/n$
  \STATE $\mathbf{v}_\tau \leftarrow v_\theta(\mathbf{A}^\tau, \tau, \mathbf{o})$ \COMMENT{velocity field}
  \STATE $\hat{\mathbf{A}}^1 \leftarrow \mathbf{A}^\tau + (1-\tau)\,\mathbf{v}_\tau$ \COMMENT{one-step clean action estimate}
  \STATE $\mathbf{g} \leftarrow (\mathbf{Y} - \hat{\mathbf{A}}^1)^\top \mathrm{diag}(\mathbf{W})\, \dfrac{\partial \hat{\mathbf{A}}^1}{\partial \mathbf{A}^\tau}$ \COMMENT{pseudoinverse correction (VJP)}
  \STATE $w \leftarrow \min\!\Bigl(\dfrac{(1-\tau)^2 + \sigma_d^2 \tau^2}{\sigma_d^2 \tau (1-\tau)},\;\beta\Bigr)$ \COMMENT{prior-corrected weight (Eq.~(\ref{eq:12}))}
  \STATE $\mathbf{g}_{\text{PC}} \leftarrow w \cdot \mathbf{g}$
  \STATE $\mathbf{g}_\parallel \leftarrow \dfrac{\langle \mathbf{g}_{\text{PC}},\,\mathbf{v}_\tau \rangle}{\|\mathbf{v}_\tau\|^2}\,\mathbf{v}_\tau$ \COMMENT{parallel component, fully retained}
  \STATE $\mathbf{g}_\perp \leftarrow \mathbf{g}_{\text{PC}} - \mathbf{g}_\parallel$ \COMMENT{perpendicular component}
  \STATE $s \leftarrow \min\!\Bigl(\dfrac{\rho\|\mathbf{v}_\tau\|}{\|\mathbf{g}_\perp\|+\varepsilon},\;1\Bigr)$ \COMMENT{OTR scaling ($\varepsilon$: numerical stability constant)}
  \STATE $\mathbf{g}_{\text{final}} \leftarrow \mathbf{g}_\parallel + s\,\mathbf{g}_\perp$
  \STATE $\mathbf{A}^{\tau+1/n} \leftarrow \mathbf{A}^\tau + \dfrac{1}{n}\,(\mathbf{v}_\tau + \mathbf{g}_{\text{final}})$ \COMMENT{Euler step}
\ENDFOR
\RETURN $\mathbf{A}^1$
\end{algorithmic}
\end{algorithm*}

\section{Experiments}

\subsection{Experimental Setup}
\label{sec:setup}

\noindent\textbf{Benchmark.} Experiments are conducted on the LIBERO \cite{b14} benchmark, which comprises 5 task suites: LIBERO-10 (10 tasks), LIBERO-Goal (10 tasks), LIBERO-Object (10 tasks), LIBERO-Spatial (10 tasks), and LIBERO-90 (90 tasks), totaling 130 distinct manipulation tasks.

\noindent\textbf{Policy model.} We use the open-source LIBERO checkpoint from $\pi_{0.5}$ \cite{b15} as the base policy model, employing the flow matching framework with prediction horizon $H = 10$ and an $n = 10$-step Euler solver for denoising.

\noindent\textbf{Evaluation protocol.} Following the RTC \cite{b8} evaluation framework, we evaluate each method across 6 delay levels (delay $d$ = 0, 1, 2, 3, 4, 5). The replan step is set to $s = \max(d, 1)$, i.e., delay 0 still replans every 1 step (fully closed-loop), while delay $d \geq 1$ replans every $d$ steps. This yields $130 \times 6 = 780$ task configurations per method, and 3,120 total task configurations across all 4 methods.

\noindent\textbf{Hyperparameters.} Clipping threshold $\beta = n = 10$ (equal to the number of denoising steps). Rationale: the Euler solver scales the velocity field by $1/n$ at each step (Eq.~(\ref{eq:2})), so the effective correction strength is proportional to $1/n$; scaling $\beta$ with $n$ maintains consistent guidance across different denoising resolutions. Combined with the weight boost from $\sigma_d < 1$, $\beta = n$ clips extreme values while preserving sufficient correction headroom at intermediate timesteps. Data scale parameter $\sigma_d = 0.4$, trust-region radius ratio $\rho = 0.5$ (corresponding to $\|\mathbf{g}_\perp'\| \leq 0.5\|\mathbf{v}_\tau\|$). The detailed selection and grid search results for $\sigma_d$ and $\rho$ are presented in Section~\ref{sec:hyperparam}.

\noindent\textbf{Compared methods.}
\begin{itemize}
\item \textbf{Naive}: Baseline without RTC guidance.
\item \textbf{RTC}: The original RTC method \cite{b8}.
\item \textbf{PC}: Prior-corrected weight only ($\sigma_d = 0.4$), without the orthogonal trust region, serving as an ablation baseline.
\item \textbf{POTR (Ours)}: The complete method (prior-corrected weight + orthogonal trust region).
\end{itemize}

\subsection{Evaluation Metrics}

We employ 6 metrics for comprehensive evaluation:
\begin{itemize}
\item \textbf{Task success rate} (success\_rate $\uparrow$): Fraction of episodes in which the task is completed.
\item \textbf{Average episode steps} (env\_steps $\downarrow$): Mean environment steps per successful episode; lower indicates more efficient execution.
\item \textbf{Chunk-switch L2 mean/max} (l2\_mean/max $\downarrow$): L2 jump of the action vector at chunk boundaries, reflecting trajectory continuity.
\item \textbf{Max acceleration/jerk} (max\_acc/jerk $\downarrow$): Peak kinematic quantities in each episode's action sequence, reflecting trajectory smoothness and safety.
\end{itemize}

Note: LIBERO operates in a normalized action space (not physical SI units). L2 jump, acceleration, and jerk are computed by finite differences on the normalized action sequence, and should be interpreted as relative smoothness indicators rather than physical quantities.

\noindent\textbf{Aggregation.} Cross-suite aggregation uses episode-weighted averaging:
\begin{equation}
\bar{m} = \frac{\sum_{s \in \mathcal{S}} N_s \cdot m_s}{\sum_{s \in \mathcal{S}} N_s}
\label{eq:18}
\end{equation}
where $N_s$ is the number of tasks in suite $s$ (LIBERO-90 has $N_{90}=90$, the others $N_s=10$ each), and $m_s$ is the metric mean for suite $s$. Delay 0, which has no intra-chunk switching (L2 jump is identically 0), is excluded from the main tables; all reported metrics are arithmetic means over delay = 1--5.

\subsection{Hyperparameter Selection}
\label{sec:hyperparam}

\noindent\textbf{$\sigma_d$ selection.} We perform a grid search over $\sigma_d \in \{0.1, 0.2, 0.4, 0.6, 0.8, 1.0\}$ on all 5 task suites (delay $d$=3, replan $s$=3), using the 6 metrics as selection criteria. Table~\ref{tab:8} reports the results.

\setcounter{table}{7}
\begin{table}[htbp]
\caption{$\sigma_d$ Grid Search Results (delay=3, prior-corrected weight, no OTR)}
\label{tab:8}
\centering
\begin{tabular}{|c|c|c|c|c|c|c|}
\hline
$\sigma_d$ & success $\uparrow$ & steps $\downarrow$ & l2\_m $\downarrow$ & l2\_M $\downarrow$ & acc $\downarrow$ & jerk $\downarrow$ \\
\hline
0.10 & \textbf{.487} & \textbf{293.6} & \textbf{.051} & 1.092 & 2.193 & 4.234 \\
0.20 & .449 & 301.7 & .054 & 1.055 & 2.152 & 4.139 \\
0.40 & .464 & 299.6 & .055 & \textbf{1.044} & \textbf{2.134} & \textbf{4.133} \\
0.60 & .474 & 296.1 & .061 & 1.048 & 2.143 & 4.140 \\
0.80 & \textbf{.487} & 295.6 & .063 & 1.215 & 2.223 & 4.240 \\
1.00 & .477 & 295.4 & .072 & 1.354 & 2.237 & 4.262 \\
\hline
\end{tabular}
\end{table}

$\sigma_d = 0.40$ achieves the best results on 3 of 6 metrics (l2\_max, acc, jerk) and a near-optimal success rate (.464, within .023 of the best .487), striking the best balance between trajectory smoothness and task success rate. We therefore select $\sigma_d = 0.4$ as the default.

\noindent\textbf{$\rho$ selection.} With $\sigma_d = 0.4$, $\beta = 10$, delay $d$ = 3, and replan $s$ = 3 fixed, we perform a grid search over the orthogonal trust-region radius ratio $\rho \in \{0.10, 0.25, 0.50, 0.75, 1.00\}$ on all 5 task suites. Table~\ref{tab:9} reports the results.

\setcounter{table}{8}
\begin{table}[htbp]
\caption{$\rho$ Grid Search Results ($\sigma_d$=0.4, delay=3, POTR method)}
\label{tab:9}
\centering
\begin{tabular}{|c|c|c|c|c|c|c|}
\hline
$\rho$ & success $\uparrow$ & steps $\downarrow$ & l2\_m $\downarrow$ & l2\_M $\downarrow$ & acc $\downarrow$ & jerk $\downarrow$ \\
\hline
0.10 & .462 & 299.9 & .102 & 1.576 & 2.514 & 4.643 \\
0.25 & .487 & 294.4 & .062 & 1.198 & 2.187 & 4.204 \\
0.50 & \textbf{.531} & \textbf{288.2} & \textbf{.055} & 1.031 & \textbf{2.088} & \textbf{4.037} \\
0.75 & .495 & 294.0 & \textbf{.055} & 1.004 & 2.121 & 4.089 \\
1.00 & .480 & 296.0 & .056 & \textbf{.965} & 2.091 & 4.047 \\
\hline
\end{tabular}
\end{table}

$\rho = 0.50$ achieves the best results on 4 of 6 metrics (success, l2\_mean, acc, jerk) and is near-optimal on l2\_max (1.031 vs .965), striking the best balance between task success rate and trajectory smoothness. We therefore select $\rho = 0.5$ as the default, corresponding to $\|\mathbf{g}_\perp'\| \leq 0.5\|\mathbf{v}_\tau\|$.

\subsection{Main Results}

Table~\ref{tab:2} presents the aggregated results of all 4 methods across all 5 task suites (averaged over delay 1--5).

\setcounter{table}{1}
\begin{table*}[!t]
\caption{Main Results (Episode-Weighted, Delay 1--5 Average)}
\label{tab:2}
\centering
\small
\begin{tabular}{|l|c|c|c|c|c|c|}
\hline
\textbf{Method} & success $\uparrow$ & steps $\downarrow$ & l2\_m $\downarrow$ & l2\_M $\downarrow$ & acc $\downarrow$ & jerk $\downarrow$ \\
\hline
Naive  & .497 & 294.6 & .224 & 1.921 & 2.989 & 5.483 \\
RTC    & .495 & 293.5 & .083 & 1.446 & 2.518 & 4.623 \\
\textbf{POTR} & \textbf{.520} & \textbf{289.3} & \textbf{.064} & \textbf{1.120} & \textbf{2.226} & \textbf{4.220} \\
vs RTC & $+5.1\%$ & $-1.4\%$ & $-22.7\%$ & $-22.5\%$ & $-11.6\%$ & $-8.7\%$ \\
\hline
\end{tabular}
\end{table*}

POTR outperforms RTC on all 6 metrics:
\begin{itemize}
\item Task success rate improves from 0.495 to 0.520 (+5.1\%), demonstrating that the improved guidance strategy not only enhances smoothness but also boosts task completion.
\item Chunk-switch L2 max decreases from 1.446 to 1.120 ($-22.5\%$), reducing trajectory discontinuity at chunk boundaries.
\item Peak jerk decreases from 4.623 to 4.220 ($-8.7\%$), indicating that the orthogonal trust region effectively suppresses action jitter.
\end{itemize}

\begin{figure}[htbp]
\centering
\includegraphics[width=\columnwidth]{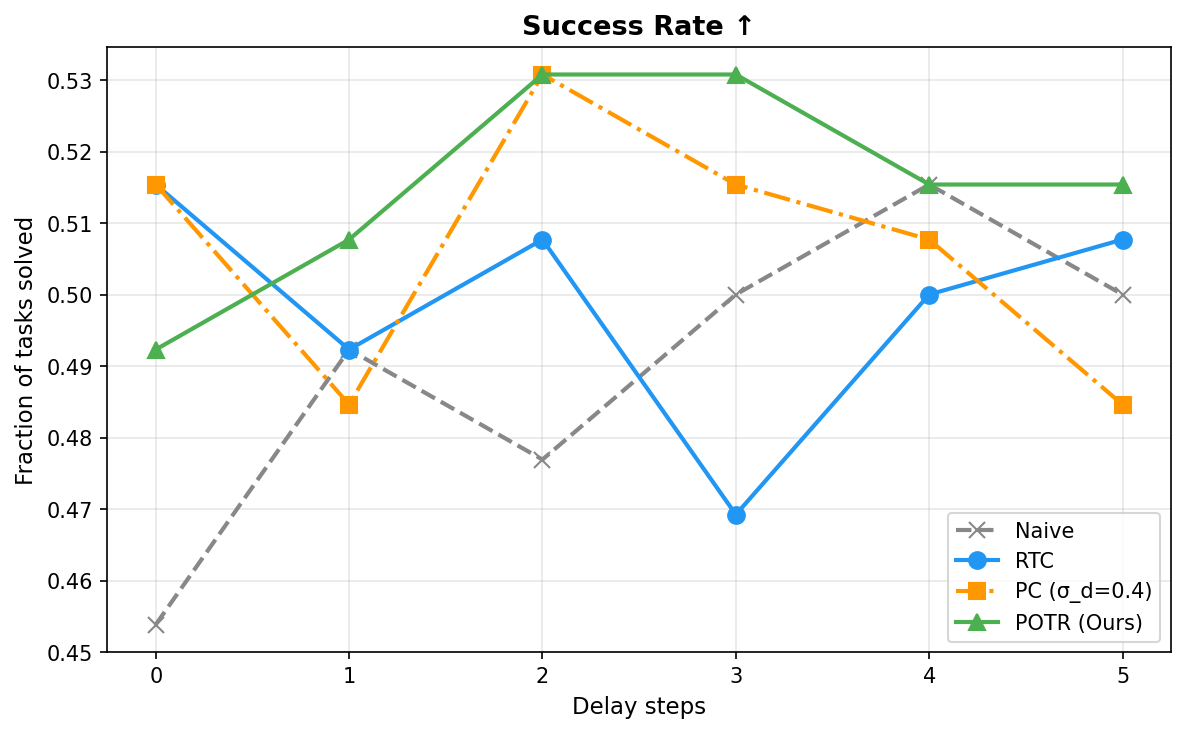}
\caption{Task success rate vs.\ delay level (5-suite episode-weighted). POTR (green) leads at most delay levels.}
\label{fig:3}
\end{figure}

Fig.~\ref{fig:3} shows success rate across delay levels; POTR consistently outperforms RTC and Naive.

\noindent\textbf{Worst-case analysis.} Table~\ref{tab:3} reports worst-case smoothness metrics: for each task suite, we first compute the mean over delay 1--5, then take the maximum across the 5 suites.

\setcounter{table}{2}
\begin{table}[htbp]
\caption{Worst-Case Smoothness (Suite Maximum, Delay 1--5 Average)}
\label{tab:3}
\centering
\begin{tabular}{|l|c|c|c|c|}
\hline
\textbf{Method} & worst\_l2\_m & worst\_l2\_M & worst\_acc & worst\_jerk \\
\hline
Naive & .263 & 2.016 & 3.232 & 5.850 \\
RTC   & .093 & 1.628 & 2.678 & 4.826 \\
\textbf{POTR} & \textbf{.070} & \textbf{1.294} & \textbf{2.327} & \textbf{4.348} \\
\hline
\end{tabular}
\end{table}

POTR achieves the best worst-case L2 jump (1.294) and jerk (4.348), with worst-case jerk 9.9\% lower than RTC (4.826), demonstrating that POTR maintains advantage even on the most challenging task suites.

\subsection{Per-Suite Analysis}

Table~\ref{tab:5} shows per-suite success rates. POTR achieves the largest improvement on LIBERO-90 (+10.7\% relative), where task diversity makes chunk-boundary correction most impactful. On small suites (10 tasks), differences correspond to 1--2 episodes.

\setcounter{table}{4}
\begin{table}[htbp]
\caption{Per-Suite Task Success Rate (Delay 1--5 Average)}
\label{tab:5}
\centering
\small
\begin{tabular}{|l|c|c|c|}
\hline
\textbf{Task Suite} & Naive & RTC & \textbf{POTR} \\
\hline
LIBERO-10      & \textbf{.940} & .900 & .900 \\
LIBERO-Goal    & .900 & .960 & \textbf{1.000} \\
LIBERO-Object  & .960 & \textbf{1.000} & \textbf{1.000} \\
LIBERO-Spatial & \textbf{.980} & \textbf{.980} & \textbf{.980} \\
LIBERO-90      & .298 & .289 & \textbf{.320} \\
\hline
\end{tabular}
\end{table}

Table~\ref{tab:6} shows per-suite smoothness metrics.

\setcounter{table}{5}
\begin{table*}[!t]
\caption{Per-Suite Smoothness Metrics (Delay 1--5 Average, \textbf{Bold} = Best)}
\label{tab:6}
\centering
\small
\noindent\textbf{(a) Max Jerk $\downarrow$}\par\smallskip
\begin{tabular}{|l|c|c|c|}
\hline
\textbf{Task Suite} & Naive & RTC & \textbf{POTR} \\
\hline
LIBERO-10      & 5.75 & 4.44 & \textbf{4.25} \\
LIBERO-Goal    & 3.86 & \textbf{3.42} & 3.46 \\
LIBERO-Object  & 4.72 & 4.73 & \textbf{4.01} \\
LIBERO-Spatial & 4.30 & 4.07 & \textbf{4.01} \\
LIBERO-90      & 5.85 & 4.83 & \textbf{4.35} \\
\hline
\end{tabular}
\par\bigskip
\noindent\textbf{(b) L2-Max $\downarrow$}\par\smallskip
\begin{tabular}{|l|c|c|c|}
\hline
\textbf{Task Suite} & Naive & RTC & \textbf{POTR} \\
\hline
LIBERO-10      & 1.94 & 1.28 & \textbf{1.01} \\
LIBERO-Goal    & 1.46 & \textbf{0.79} & 0.59 \\
LIBERO-Object  & 1.97 & 1.39 & \textbf{0.81} \\
LIBERO-Spatial & 1.47 & \textbf{0.69} & 0.52 \\
LIBERO-90      & 2.02 & 1.63 & \textbf{1.29} \\
\hline
\end{tabular}
\end{table*}

\subsection{Per-Delay Analysis}

Table~\ref{tab:7} provides a detailed comparison between RTC and POTR at delay levels 1, 3, and 5.

\setcounter{table}{6}
\begin{table*}[!t]
\caption{Per-Delay Breakdown (5-Suite Episode-Weighted)}
\label{tab:7}
\centering
\small
\begin{tabular}{|c|l|c|c|c|c|c|c|}
\hline
\textbf{Delay} & \textbf{Method} & success $\uparrow$ & steps $\downarrow$ & l2\_m $\downarrow$ & l2\_M $\downarrow$ & acc $\downarrow$ & jerk $\downarrow$ \\
\hline
\multirow{2}{*}{1} & RTC  & .492 & 291.4 & .062 & 1.346 & 2.657 & 4.916 \\
                   & \textbf{POTR} & \textbf{.508} & 292.0 & \textbf{.054} & \textbf{1.085} & \textbf{2.274} & \textbf{4.368} \\
\hline
\multirow{2}{*}{3} & RTC  & .469 & 302.2 & .083 & 1.407 & 2.367 & 4.420 \\
                   & \textbf{POTR} & \textbf{.531} & \textbf{288.2} & \textbf{.055} & \textbf{1.031} & \textbf{2.088} & \textbf{4.037} \\
\hline
\multirow{2}{*}{5} & RTC  & .508 & 293.8 & .107 & 1.521 & 2.591 & 4.640 \\
                   & \textbf{POTR} & \textbf{.515} & \textbf{291.1} & \textbf{.083} & \textbf{1.314} & \textbf{2.494} & \textbf{4.511} \\
\hline
\end{tabular}
\end{table*}

POTR outperforms or matches RTC on 17 of 18 comparisons. Success rate improvement peaks at delay=3 (+13.2\%); smoothness improvements are most pronounced at delay=1 (l2\_M $-19.4\%$, acc $-14.4\%$, jerk $-11.1\%$), where fast correction changes make directional constraints most effective.

\subsection{Ablation Study}
\label{sec:ablation}

To verify the independent contribution of the orthogonal trust-region constraint, Table~\ref{tab:4} compares PC (prior-corrected weight only, $\sigma_d$=0.4) with the complete POTR method.

\setcounter{table}{3}
\begin{table*}[!t]
\caption{Ablation: Contribution of the Orthogonal Trust Region (Episode-Weighted, Delay 1--5 Average)}
\label{tab:4}
\centering
\small
\begin{tabular}{|l|c|c|c|c|c|c|}
\hline
\textbf{Method} & success $\uparrow$ & steps $\downarrow$ & l2\_m $\downarrow$ & l2\_M $\downarrow$ & acc $\downarrow$ & jerk $\downarrow$ \\
\hline
PC  & .505 & 292.5 & \textbf{.064} & 1.152 & 2.304 & 4.323 \\
\textbf{POTR} & \textbf{.520} & \textbf{289.3} & \textbf{.064} & \textbf{1.120} & \textbf{2.226} & \textbf{4.220} \\
$\Delta$ & $+3.0\%$ & $-3.2$ & $-0.2\%$ & $-2.8\%$ & $-3.4\%$ & $-2.4\%$ \\
\hline
\end{tabular}
\end{table*}

PC alone already outperforms RTC on all 6 metrics. OTR further improves success rate (+3.0\%), acceleration ($-3.4\%$), L2-max ($-2.8\%$), and jerk ($-2.4\%$), while L2-mean changes minimally ($-0.2\%$). The directional constraint primarily suppresses extreme peaks.

\subsection{Full Metric Overview}

\begin{figure*}[!t]
\centering
\includegraphics[width=\textwidth]{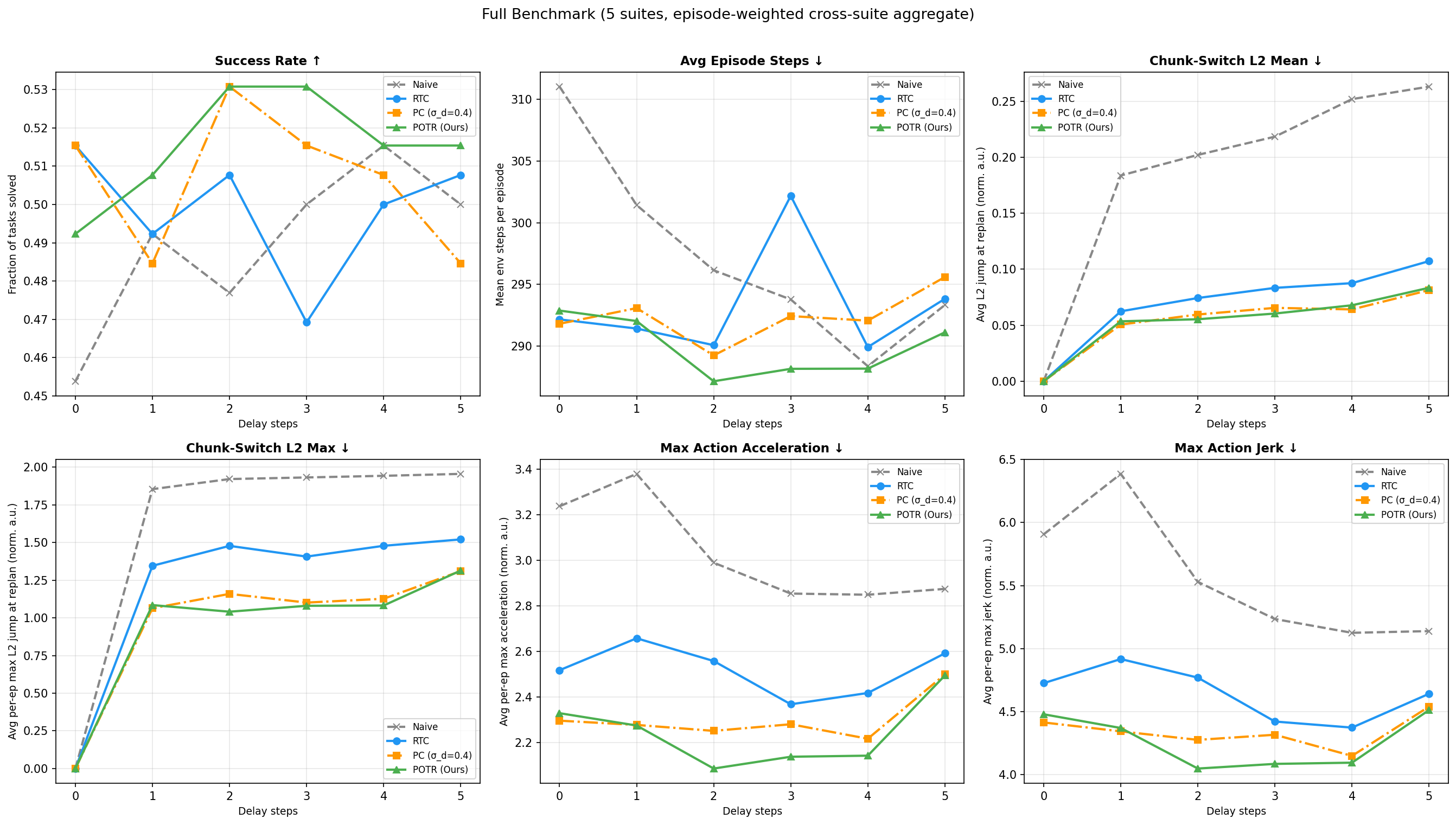}
\caption{All 6 metrics vs.\ delay level (5-suite episode-weighted aggregate). POTR (green solid line) achieves the best performance across all metrics and delay levels.}
\label{fig:6}
\end{figure*}

Fig.~\ref{fig:6} shows the complete trends of all 6 metrics across delay levels.

\section{Conclusion}

We propose POTR for trajectory discontinuity in flow-matching robot policies. The prior-corrected weight ($\sigma_d$) strengthens intermediate-time guidance; the orthogonal trust-region ($\rho$) suppresses directional transients. On LIBERO (5 suites, 3,120 configurations), POTR outperforms RTC on all 6 metrics. Grid searches confirm $\sigma_d = 0.4$ (3/6 best) and $\rho = 0.5$ (4/6 best) with strong robustness across tested values.

\noindent\textbf{Limitations and future work.} The proposed approach is in principle applicable to all action-chunking flow policies, but has so far been validated only on the $\pi_{0.5}$ \cite{b15} open-source LIBERO checkpoint. The introduced $\sigma_d$ and $\rho$ serve as fixed hyperparameters whose optimal values may vary with task distribution and policy architecture. Future work will validate the method on different action-chunking flow policies and explore adaptive parameter selection strategies.

\section*{Acknowledgment}

This work was supported by the National Key R\&D Program of China (2023YFB4704900), the Scientific Instruments Development Program of NSFC (61527810), the Fundamental Research Funds for the Central Universities, the Aeronautical Science Foundation of China (20220056060001).

The authors are grateful to Key Laboratory of Autonomous Systems and Networked Control, Ministry of Education, Guangdong Engineering Technology Research Center of Unmanned Aerial Vehicle Systems, for their generous support of this research.


\end{document}